\pdfoutput=1

\documentclass[11pt]{article}

\usepackage[]{acl}

\usepackage{times}
\usepackage{latexsym}
\usepackage{algorithm}
\usepackage{algorithmic}

\usepackage{microtype}

\usepackage{amsmath}
\usepackage{booktabs}
\usepackage{makecell}
\usepackage{multirow}
\usepackage{pifont}
\usepackage{multicol}
\usepackage{amssymb}
\usepackage{latexsym}
\usepackage{graphicx}
\usepackage{ulem}
\usepackage{tikz}
\usepackage{pgfplots}
\usepackage{tikz-dependency}
\usepackage{subcaption}

\usepackage{microtype}
\usepackage[switch]{lineno}

\usepackage{newfloat}
\usepackage{listings}
\floatstyle{ruled}
\newfloat{listing}{tb}{lst}{}
\floatname{listing}{Listing}

\usepackage[T1]{fontenc}

\usepackage[utf8]{inputenc}

\usepackage{microtype}

{}
{}
{}
{}
{}
{}
{}
{}

\title{Pro-KD: Progressive Distillation by Following the Footsteps of the Teacher}

\author{Mehdi Rezagholizadeh$^{1}$\thanks{\ \ Equal Contribution}  \quad Aref Jafari$^{2,1*}$ \quad Puneeth Saladi$^1$ \quad \\ \textbf{Pranav Sharma$^2$} \quad \textbf{Ali Saheb Pasand$^{1,2}$} \quad \textbf{Ali Ghodsi$^{2}$} \\
  $^1$Huawei Noah's Ark Lab \\
  $^2$University of Waterloo \\
  \small \texttt{\{mehdi.rezagholizadeh,aref.jafari,puneeth.saladi\}@huawei.com} \\

}

\begin{document}
\maketitle

\begin{abstract}
With ever growing scale of neural models, knowledge distillation (KD) attracts more attention as a prominent tool for neural model compression.
However, there are counter intuitive observations in the literature showing some challenging limitations of KD. A case in point is that the best performing checkpoint of the teacher might not necessarily be the best teacher for training the student in KD. Therefore, one important question would be how to find the best checkpoint of the teacher for distillation? Searching through the checkpoints of the teacher would be a very tedious and computationally expensive process, which we refer to as the \textit{checkpoint-search problem}. Moreover, another observation is that larger teachers might not necessarily be better teachers in KD which is referred to as the \textit{capacity-gap} problem. 
To address these challenging problems, in this work, we introduce our progressive knowledge distillation (Pro-KD) technique which defines a smoother training path for the student by following the training footprints of the teacher instead of solely relying on distilling from a single mature fully-trained teacher. 
We demonstrate that our technique is quite effective in mitigating the capacity-gap problem and the checkpoint search problem. 
We evaluate our technique using a comprehensive set of experiments on different tasks such as image classification (CIFAR-10 and CIFAR-100), natural language understanding tasks of the GLUE benchmark, and question answering (SQuAD 1.1 and 2.0) using BERT-based models and consistently got superior results over state-of-the-art techniques.
\end{abstract}

\begin{table}[t]
\caption{The best performing checkpoints vary for different size of the models, and different tasks. It is evident that the best checkpoint of the teacher does not necessarily lead to the best performing student model.} 
\resizebox{\columnwidth}{!}{%
\begin{tabular}{c c c c c} 
\Xhline{5\arrayrulewidth} 
Task & Model & \makecell{Best \#ChPt \\ Epoch } & ACC  & $\Delta \uparrow$ \\[0.5ex]
\hline 
\multirow{3}{*}{MRPC}& BERT$_{\text{LARGE}}$($\spadesuit$T) & \textbf{ 6} & 88.22& -\\ 
& BERT$_{\text{SMALL}}$($\clubsuit$S) & \textbf{ 3} & 85.29 & \textbf{+0.7}\\
& DistilBERT($\clubsuit$S) & \textbf{ 5} & 89.16 & \textbf{+0.8}\\
\hline
\multirow{3}{*}{SST-2}& BERT$_{\text{LARGE}}$($\spadesuit$T) & \textbf{ 1} & 92.89 & -\\ 
& BERT$_{\text{SMALL}}$($\clubsuit$S) & \textbf{ 7} & 88.76 & \textbf{+0.6}\\
& DistilBERT($\clubsuit$S) & \textbf{ 7} & 91.63 & \textbf{+0.4}\\
\hline 
\multirow{3}{*}{QNLI}& BERT$_{\text{LARGE}}$($\spadesuit$T) & \textbf{ 6} & 92.4& -\\ 
& BERT$_{\text{SMALL}}$($\clubsuit$S) & \textbf{ 2} & 87.08 & \textbf{+0.3}\\
& DistilBERT($\clubsuit$S) & \textbf{ 4} & 90.46& \textbf{+0.2}\\
\hline
\end{tabular}
}
\caption*{\small $\spadesuit$T:Teacher, $\clubsuit$S: Student, \#ChPt: Checkpoint Number, ACC: Accuracy, $\Delta \uparrow$: Accuracy improvement compared to the best checkpoint of the teacher }
\label{table:search} 
\end{table}

\section{Introduction}
Knowledge distillation (KD)~\cite{hinton2015distilling} has gained a lot of attention in different deep learning applications such as natural language processing (NLP)~\cite{sunmobilebert,sun2019patient, jiao2019tinybert,clark2019bam}, computer vision~\cite{guo2020online, mirzadeh2019improved}, and speech processing~\cite{essence,tutornet,distill}. 
Nowadays, the scale of neural networks is growing in the favor of improving their performance~\cite{devlin-etal-2019-bert}. A case in point is pre-trained language models (PLMs) such as the GPT-3~\cite{brown2020language}, Pangu-$\alpha$~\cite{zeng2021pangu}, and WuDao2 which have more than a hundred billions of parameters~\cite{brown2020language}. However, deploying these models on devices with limited computational power will be very challenging, if not impossible. In this regard, KD can be used as one of the most prominent neural model compression techniques.

KD adds a new loss term to the regular cross-entropy classification loss. This new loss encourages the student model to mimic the output of a pre-trained teacher network. The teacher network is usually a higher capacity model which is able to learn the underlying function of the training data to a good extent. The output prediction of the teacher is called \textit{soft-target} for the student model. In contrast to ground-truth labels coming from the training dataset which only carry the information about a single class, soft-targets can provide some information about the relative distribution of different classes for each training data. Therefore, a pre-trained teacher is able to provide some auxiliary signal besides the labels in the training dataset.  

KD has been investigated a lot in the literature. \citet{sun2019patient,passban2020alpkd,wu-etal-2020-skip} proposed a technique to improve KD by incorporating the intermediate layer matching in the KD loss. \citet{jiao2019tinybert} show a two-stage KD with intermediate layer mapping, attention distillation and embedding distillation for BERT-based models. Mate-KD~\cite{rashid-etal-2021-mate} and MiniMax-KNN KD~\cite{kamalloo-etal-2021-far} tailor data augmentation for KD, in which augmented samples are generated or selected based on maximum divergence loss between the student and teacher networks. \citet{rashid2020towards} propose a zero-shot KD technique in NLP in which the student does not need to access the teacher training data for its training.   
\citet{clark2019bam} use KD for multi-task learning in natural language understanding. \citet{kim2016sequence} propose a sequence-level KD solution for machine translation. 
\citet{guo2020online} introduces a collaborative training of students with different capacities with KD.

Although KD has been successful in many different deep learning tasks, it is subject to some special limitations as well.
For example, it is shown in~\cite{lopez2015unifying} that, based on VC-dimension, the teacher capacity should not be too large in KD. 
Similar observation is given by~\citet{mirzadeh2019improved} using empirical and theoretical justifications that KD will be less effective when the capacity gap between the teacher and student is large. This problem is referred to as the \textit{capacity-gap} problem. \citet{mirzadeh2019improved} proposed the TA-KD solution to this problem by adding one or multiple intermediate teacher assistant (TA) networks to learn from the teacher using KD and then train the student using other distillation processes. 
However, training intermediate networks can be prohibitive in terms of adding to the training time, computational complexity and extra error propagation.   
The other case in point is that a fully-trained teacher might not be the best teacher for the student~\cite{cho2019efficacy}. In other words, an early-stopped teacher can be a better option for KD compared to a fully-trained one.    
This observation implies that we need to search through the checkpoints of the teacher to find the best model for the distillation process. However, this search can be very expensive especially when we deal with PLMs. We refer to this problem as the \textit{checkpoint-search} problem. Our investigations of this checkpoint search problem shows the significance of this issue especially dealing with PLMs. Table~\ref{table:search} depicts that the best distillation checkpoint of the teacher varies for each task and for each student configuration and it is very different from the teacher best performing checkpoint.

In this work, we propose our Pro-KD solution to tackle both of the capacity-gap and checkpoint-search problems. 
In Pro-KD, the student grows gradually with the growth of the teacher. We hypothesize that the training path of the teacher can be informative for the student and we should not disregard it. Therefore, in contrast to the original KD where the student learns from the best pre-trained teacher, in Pro-KD the student starts its learning process together with the teacher. Furthermore, in contrast to the TA-KD technique which reduces the capacity of the teacher by adding intermediate TAs to the distillation process, 
in our Pro-KD, we mitigate the capacity gap by making the training path of the student more smooth and gradual by following the training footsteps of the teacher. Moreover, to make the training smooth further for the student, inspired by~\cite{jafari-etal-2021-annealing}, we apply an adaptive temperature factor to the output of the teacher while being trained. This temperature factor is decreased during the training. We will show the effectiveness of our solution using theoretical justification and empirical evaluations. We evaluate Pro-KD by performing experiments on both NLP (the GLUE benchmark and SQuAD 1.1 and 2.0) and image classification tasks (CIFAR-10, CIFAR-100). The contributions of this paper are summarized in the following: 
\begin{enumerate}
    \item We propose our Pro-KD solution to the KD \textit{capacity-gap} and \textit{checkpoin-search} problems. In Pro-KD, the student follows the training footsteps of the teacher. Moreover, it introduces a dynamic temperature function to the output of the teacher when it is distilled to the student to make the student training gradual and smooth.
    \item We apply our technique to ResNET8 model on both CIFAR-10 and CIFAR-100 image classification tasks, and the natural language inference task on different BERT based models such as DistilRoBERTa, and BERT-Small on the GLUE benchmark and also SQuAQ 1.1 and 2.0 question answering datasets and achieved the state-of-the-art results.  
    \item Our technique is simple, architecture agnostic, does not require training any extra network and can be applied on top of different variants of KD.
\end{enumerate}

\section{Background }
\paragraph{Knowledge Distillation} KD~\cite{hinton2015distilling} is a well-known method for neural model compression and also is shown to be an effective regularizer in improving the performance of neural networks in the self-distillation~\cite{yun2020regularizing,hahn2019self} or born-again~\cite{furlanello2018born} setups. KD adds a particular loss term to the regular cross entropy (CE) classification loss: 
\begin{equation}
\begin{split}
    & \mathcal{L}_{KD} (\phi) = CE\Big( y, S(x;\phi)\Big)+\\ 
    & \mathcal{T}^2KL\Big(\sigma(\frac{z_t(x;\theta)}{\mathcal{T}}),\sigma(\frac{z_s(x;\phi)}{\mathcal{T}})\Big) \\
\end{split}
\label{eq:T}
\end{equation}
where $x$ is the input data and $y$ is its associated label, $\phi$ and $\theta$ refer to the student and teacher parameters, $\sigma$ is the softmax function, $z_s$ and $z_t$ are the student and teacher logits, $\mathcal{T}$ is the temperature parameter to control the softness of the output probability distributions, $CE$ and $KL$ refer to the cross entropy and KL divergence loss functions respectively.    

Regular KD training is a two-stage process in which the teacher is fully trained in the first stage and deployed in training of the student model in the next stage. The student is trained based on the \textit{hard labels} coming from the ground-truth training data and \textit{soft labels} coming from the teacher output predictions. 

\section{Related Work}
In this section, we review the most related works to our paper in the literature. 


\subsection{Capacity-Gap Problem in KD}

The capacity-gap problem refers to having a more powerful or larger teacher in KD is not necessarily going to give rise to training a better student. \citet{mirzadeh2019improved} propose their TA-KD solution to this problem by introducing an intermediate TA network whose capacity is greater than the student but smaller than the teacher. The target of this TA network is to learn from the teacher and train the student. 
It is evident that in this setup the capacity gap between the TA and student is less than that of the main teacher and the student. Even though adding the TA network can mitigate the capacity-gap problem, there are two downsides in this technique: first, training a separate TA network is costly; second, the sequence of training multiple networks can lead to error accumulation and error propagation to the student. Moreover, TA-KD only showed their results on image classification tasks. 
\citet{jafari-etal-2021-annealing} proposed a TA-free solution for the capacity-gap problem which is called Annealing-KD. Annealing-KD adds an annealed dynamic temperature factor to the output of the teacher to make the training process for the student very gradual. The temperature factor starts from a large value to apply the maximum smoothing to the output of the teacher at the beginning of the training and the gradually tends toward 1 where there is no smoothing effect on the output of the teacher. Our Pro-KD is inspired by Annealing-KD in using a dynamic temperature factor to solve the capacity-gap problem, but we can highlight two main improvements over Annealing-KD: first, Annealing-KD suffers from the checkpoint search problem while Pro-KD does not; second, Pro-KD takes advantage of the intrinsic gradual training of the teacher training which is not considered in Annealing-KD. 


\subsection{Checkpoint Search Problem in KD}
\citet{choi-etal-2020-toward} highlighted their findings over computer vision (CV) models that bigger models are not necessarily better teachers and also early stopped teachers can train students better. To overcome these issues they propose an ad-hoc approach called Early Stopped KD (ESKD); however, ESKD still needs to search among the pool of teacher's early checkpoints and the selection mechanism among this early checkpoints is not clear. Moreover, in PLMs, generally teachers are not fine-tuned for long and yet searching among the checkpoints of large PLMs is expensive.  
\cite{jin2019knowledge} proposed Route Constrained Optimization (RCO) distillation which is a curriculum learning technique to follow easy-to-hard training scheme on top of the teacher's training trajectory. We can deem this technique as the most related work to us with the following differences: 1- our Pro-KD has a temperature factor to make the training process more gradual which can handle the capacity gap better; 2- RCO still requires search to define its trajectory; 3- we evaluated Pro-KD on both CV and NLP tasks but RCO is only evaluated on CV tasks. In this paper we have tried RCO on both NLP and CV tasks and results indicate that our Pro-KD outperforms RCO consistently.

In the next section, we introduce our Pro-KD solution to the capacity-gap and checkpoint-search problems. Pro-KD exploits a smooth training process for the student to gradually learn from the teacher while being trained from scratch, instead of learning from a fully-trained Teacher. 


\begin{figure*}[h]
    \centering
    \includegraphics[width=0.8\textwidth]{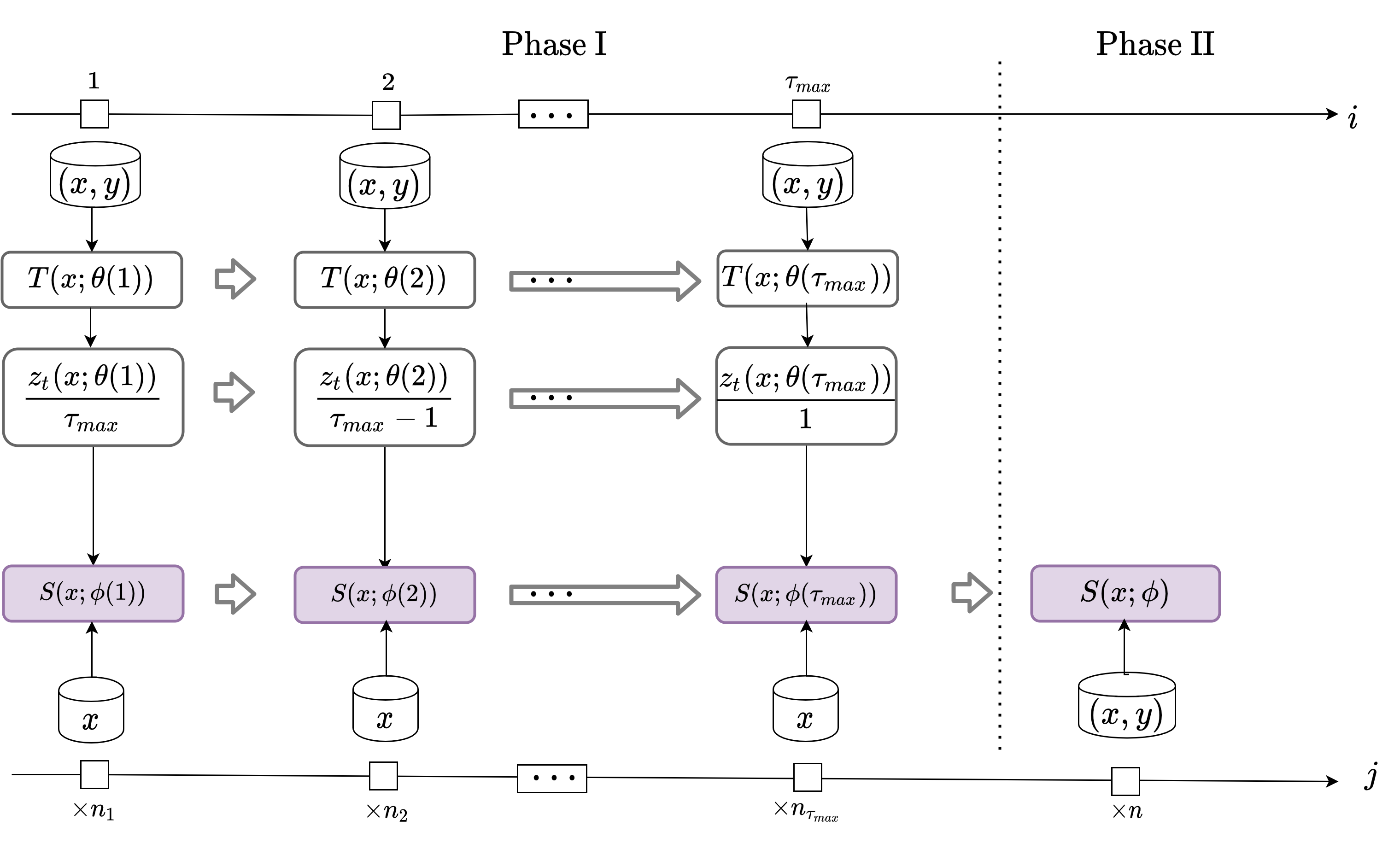}
    \caption{ Phase I and Phase II of the Pro-KD method. Phase I: the teacher is trained on the labeled training data.  The student at each step tries to mimic the behavior of the corresponding teacher checkpoint. The logits of the teacher at each time step is attenuated with the temperature parameter.  We start training of the student from $\mathcal{T}=\tau_{max}$ and go to $\mathcal{T}=1$. Phase II: training the student with the labeled data only using the cross entropy loss. $n_i$ for $1 \leq i \leq \tau_{max} $ refers to the number of training epochs of the student corresponding to the $i^\text{th}$ training epoch of the teacher. $n_i$s should add up to $N$ which is the preset total number of training epochs of the student. Bear in mind that since the teacher is usually early stopped, teacher is trained less longer than the student model. }
    \label{fig:fig1}
\end{figure*}

\section{Methodology: Pro-KD}
Methodology of this paper concerns addressing the capacity-gap and checkpoint-search problems in KD. We propose our Pro-KD technique which is a progressive training procedure in which the student is trained together with the teacher. Therefore in Pro-KD, the student is not exposed to the output of a fully trained teacher from beginning, and instead, it learns from the teacher at the same time while the the teacher is being trained. This training process for the student is more gradual and also the student can learn from the training path of the teacher. Inspired by~\citet{jafari-etal-2021-annealing}, We define the training in two phases: first, the student is only supervised by an adaptive smoothed version of the teacher; second, the student will be only trained on ground-truth labels. The detail of each phase is explained in the following.  

\paragraph{Phase I) General Step-by-Step Training with a Teacher} In this phase, the teacher is being trained using the cross entropy loss on the training data. Without lose of generality, let's assume that $\theta(i)$, the optimized parameters of the teacher at the beginning of epoch $i$, will be updated during the $i^{\text{th}}$ epoch of training to obtain $\theta(i+1)$:   

\begin{equation}
\begin{split}
    & \mathcal{L}_T (\theta(i)) =CE\big( y, T(x;\theta(i))\big) \\
    &  \theta(i+1) \leftarrow \min_{\theta} \mathcal{L}_T (\theta(i)) \\
\end{split}
\label{eq:T}
\end{equation}
Then, a smoothed version of the teacher output using a temperature factor $\mathcal{T}_i$ at epoch $i$ will be used to train the student at epoch $j$: 

\begin{equation}
\begin{split}
    & \mathcal{L}^I_S (\phi(j)) =\| z_s(x;\phi(j))-\frac{z_t(x;\theta(i))}{\mathcal{T}_i}\|_2^2 \\
    & \phi(j+1) \leftarrow \min_{\phi} \mathcal{L}_S^I (\phi(j)) \\
    & \mathcal{T}_{i+1} = \mathcal{T}_{i}-1;\text{  } \mathcal{T}_{1} = \tau_{max}  \\
\end{split}
\label{eq:KD}
\end{equation}
where $\mathcal{L}^I_S$ refers to the student loss function in phase I,  $z_s$ and $z_t$ are the logits of the student and teacher respectively, $\phi(j)$ represents the parameters of the student model at epoch $j$, $\mathcal{T}_{i}$ is adaptive temperature factor. The rational behind applying the temperature to the teacher output is to make the output of the teacher smoother for the student to learn, especially at early stages of the training process~\citep{jafari-etal-2021-annealing}.  

Considering that the teacher is usually early-stopped and it is not trained for long, the number of training epochs of the student is longer than that of the teacher. Therefore, for each given epoch $i$ of the teacher, the student can be trained for $ n_i \geq 1$ epochs, subject to $\sum_{i=1}^{\tau_{max}} n_i = N$ where $N$ is the preset total number of student training epochs. $n_i$ is set to a constant integer number in most of our experiments but can be customized as well. 

Bear in mind that we apply an adaptive temperature to the teacher logit in the student distillation loss, and this temperature starts from the highest $\tau_{max}$ value and decrease linearly with the epoch number of the teacher throughout training. $\tau_{max}$ is a hyper-parameter in our training. Moreover, we keep the temperature fixed at each epoch $i$. After training with the teacher logits, in the next phase the student model will be trained on the ground-truth labels. 

\paragraph{Phase II) Training with the Ground-Truth Labels} 
After training the student based on the soft targets of the teacher, the student is trained on the ground-truth labels using the cross entropy loss for a few epochs:

\begin{equation}
\begin{split}
    & \mathcal{L}_S^{II} (\phi(j)) =CE\big( y, S(x;\phi(j))\big) \\
    &  \phi(j+1) \leftarrow \min_{\phi} \mathcal{L}_S^{II} (\phi(j)). \\
\end{split}
\label{eq:T}
\end{equation}

In summary, our technique enables the student to learn from the teacher more smoothly. Pro-KD is different from regular KD technique in the following aspects: 
\begin{enumerate}
    \item rather than distilling from a fully trained teacher; we distill from the teacher during its training; 
    \item in contrast to regular KD which applies a fixed temperature parameter to both networks, we apply an adaptive temperature factor only to the logits of the teacher; 
    \item we apply the KD loss and the cross entropy loss in two separate phases. 
\end{enumerate}

\subsection{Why Does Pro-KD work?}

\citet{lopez2015unifying} do an analysis based on VC-dimension theory to discuss the conditions under which KD works better than no-KD scenarios.  
It is shown that KD works if the following inequality holds:  

\begin{equation}
O(\frac{|\mathcal{F}_s|_c+|\mathcal{F}_t|_c}{n^{\alpha}}) + \varepsilon_{t}+\varepsilon_{l} \leq O(\frac{|\mathcal{F}_s|_c}{\sqrt{n}}) + \varepsilon_{s}
\label{eq:vap}
\end{equation} 
where 
$\mathcal{F}_s$ and $\mathcal{F}_t$ are the function classes corresponding to the teacher and student;  $|.|_c$ is a function class capacity measure; $O(.)$ is the estimation error of training the learner; $\varepsilon_s$ is the approximation error of the best estimator function belonging to the $\mathcal{F}_s$ class with respect to the underlying function; $\varepsilon_t$ is a similar approximation error for the teacher with respect to the underlying function; $\varepsilon_l$ is the approximation error of the best student function with respect to the teacher function; $n$ is the number of training samples, and $\frac{1}{2}\leq \alpha \leq 1$ is a parameter related to the difficulty of the problem. Smaller values of $\alpha$ indicate slower training and larger values correspond to faster learning rates of the student. Given Eq.~\ref{eq:vap} we observer that for larger teacher models, the value of the left hand side of the inequality might get higher than that of the right hand side, which implies that KD might not work as expected for large capacity teachers. 


We analyzed the capacity-gap problem based on the inequality~\ref{eq:vap}. To satisfy this inequality when the capacity of the teacher is large (or equivalently the capacity gap between the two networks is large), we can think about reducing the capacity of the teacher (eg. the TA-KD technique~\cite{mirzadeh2019improved} introduces a smaller TA network instead of a large teacher), or increase $\alpha$. Increasing $\alpha$ is equivalent to making the training process of student easier and smoother.   
It has been shown in~\cite{li2017visualizing,chaudhari2019entropy} that learning a smoother loss is easier than a sharp one. Therefore, we tried to make the training of the student smoother by following the teacher's training steps and also applying the adaptive temperature factor. In addition to this justification, we present a theoretical justification in the Appendix.

\begin{table}[t]
\caption{Comparing the test accuracy of Pro-KD, TAKD~\citep{mirzadeh2019improved}, Annealing-KD~\citep{jafarpour-etal-2021-active}, RCO~\citep{jin2019knowledge}, regular KD, and student without teacher on CIFAR-10 dataset with both ResNet and CNN models } 
\centering 
\resizebox{\columnwidth}{!}{%
\begin{tabular}{c c c c} 
\Xhline{4\arrayrulewidth} 
Model & Type & Training method & Accuracy\\ [0.5ex] 
\hline 
\multirow{6}{*}{ResNet}& Teacher(110) & from scratch & 93.8\\ 
& TA(20) & KD & 92.39\\
\cline{2-4}
& Student(8) & from scratch & 88.44\\
& Student(8) &  KD & 88.45\\
& Student(8) & TAKD & 88.47\\
& student(8) & RCO & 88.90 \\
& Student(8) & Annealing-KD & 89.44\\
& Student(8) & \textbf{Pro-KD (ours)} & \textbf{90.01}\\
\hline 
\end{tabular}
}
\label{tableCV1:nonlin} 
\end{table}

\begin{table}[t]
\caption{Comparing the test accuracy of Pro-KD, TAKD~\cite{mirzadeh2019improved}, Annealing-KD~\citep{jafarpour-etal-2021-active}, RCO~\cite{jin2019knowledge}, regular KD, and student without teacher on CIFAR-100 dataset with both ResNet and CNN models} 
\centering 
\resizebox{\columnwidth}{!}{%
\begin{tabular}{c c c c} 
\Xhline{4\arrayrulewidth} 
Model & Type &Training method & Accuracy\\ [0.5ex] 
\hline 
\multirow{6}{*}{ResNet} & teacher(110) & from scratch & 71.92\\ 
& TA(20) &  KD & 67.6\\
\cline{2-4}
& student(8) & from scratch & 61.37\\
& student(8) &  KD & 61.41\\
& student(8) & RCO & 61.62 \\
& student(8) & TAKD & 61.82\\
& student(8) & Annealing KD & 63.1\\
& student(8) & \textbf{Pro-KD (ours)} & \textbf{63.43}\\
\hline 
\end{tabular}
}
\label{tableCV2:nonlin} 
\end{table}

\section{Experiments and Results}
In this section, we evaluate our Pro-KD on 3 different sets of experiments: on image classification, natural language understanding and question answering tasks. In all these three experiments, we compare Pro-KD with the state-of-the-art techniques such as TA-KD~\cite{mirzadeh2019improved}, Annealing-KD~\citep{jafari-etal-2021-annealing} and RCO~\citep{jin2019knowledge} technique and other baselines such as the original KD method~\cite{hinton2015distilling} and also without KD baselines. 

\subsection{Experimental Setup for Image Classification Tasks}

\paragraph{Data}
We used CIFAR-10 and CIFAR-100~\cite{krizhevsky2009learning} datasets for our experiments on the image classification tasks. Both datasets have 60,000 of $32\times 32$ color images distributed into 50,000 training and 10,000 test samples, with 10 classes for CIFAR-10 and 100 classes for CIFAR-100 

\paragraph{Setup}
For these experiments, we used ResNet-8 as the student and resNet-110 as the teacher models. The experimental setups are similar to TAKD method~\cite{mirzadeh2019improved}. Also for the TAKD baseline, we used ResNet-20 as the TA model. The results of these experiments can be found in table \ref{tableCV1:nonlin} and \ref{tableCV2:nonlin} for comparison. For the baselines, first, the ResNet-110 teacher is trained from scratch on the given datasets and then it is used for training baselines with KD. For TAKD baseline, the original KD method is applied to train the TA network with the teacher network and it is applied to train the student network with TA network. For training the student with the proposed Pro-KD method, we trained the teacher for 160 epochs. For training the student, we used maximum temperature 10 and learning rate 0.1. For every 16 epochs we decrease the temperature. Also, we used 140 epochs as warm-up epochs and the increment factor is 1.


\begin{table*}[ht]
\caption{Dev set results of training DistilRoBERTa using the RoBERTa$_\text{Large}$ model on the GLUE benchmark.} 
\small
\centering 
\begin{tabular}{c c c c c c c c c c} 
\Xhline{4\arrayrulewidth} 
KD Method & CoLA & RTE & MRPC & STS-B & SST-2 & QNLI & QQP & MNLI &  Score\\ [0.5ex] 
\hline 
RoBERTa$_\text{Large}$ (Teacher) & 67 & 85 & 91.63 & 92.53 & 96.21 & 94.53 & 91.45 & 89.94/89.97 &88.54\\ 
\hline
DistilRoBERTa(NoKD)& 61.9&	69.31&	89.85&	88.46&	91.86&	91.31&	90.04&	84.03/	83.69&	83.32 \\
Vanilla KD & 60.97 & 71.11 & 90.2 & 88.86 & 92.54 & 91.37 & 91.64 & 84.18/84.11 & 83.85\\ 
TAKD & 61.15 & 71.84 & 89.91 & 88.94 & 92.54 & 91.32 & 91.7 & 83.89/84.18  & 83.93\\
RCO&60.66 &72.2 & 90.56 & 88.41& 91.97&91.09  &90.04 &88.04/84.18 & 63.63\\  
{Annealing KD} & {61.67} & {73.64} & {90.6} & {89.01} & {93.11} & {91.64} & 91.5 & {85.34/84.6} &  {84.52}\\
\textbf{Pro-KD (Ours)} & {62.14} & {73.64} & {91.9} & {88.8} & {92.66} & {91.47} & 91.53 & {84.83/84.84} &  {\textbf{84.62}}\\
\hline 
\end{tabular}
\label{tabledistilroberta:nonlin} 
\end{table*}

\begin{table*}[ht]
\caption{Test set results of DistilRoBERTa trained on the GLUE tasks using the RoBERTa$_\text{Large}$ teacher model. } 
\small
\centering 
\begin{tabular}{c c c c c c c c c  c} 
\Xhline{4\arrayrulewidth} 
KD Method & CoLA & RTE & MRPC & STS-B & SST-2 & QNLI & QQP & MNLI  & Score\\ [0.5ex] 
\hline 
Vanilla KD & 54.3 & 74.1 & 86/80.8 & 85.7/84.9 & 93.1 & 90.8 & 71.9/89.5 & 83.6/82.9  & 80.62\\ 
TAKD & 53.2 & \textbf{74.2} & 86.7/82.7 & 85.6/84.4 & 93.2 & \textbf{91.0} & 72/89.4 & 83.8/83.2  & 80.69\\
RCO & 55.1 & 73.0 & \textbf{89.7/86.0}  & 85.0/83.8 & 93.5 & 88.9 & 69.5/87.6 & 83.2/82.4  & 80.51\\
{Annealing KD} & 54& 73.7& 88.0/83.9& 87.0/86.6& \textbf{93.6}&  90.8& \textbf{72.6/89.7}&83.8/ 83.9& 81.23 \\
\textbf{Pro-KD (Ours)} & \textbf{55.8} & {73.6} & 88.7/84.2 & \textbf{87.2/86.7} & {93.4} & \textbf{91.0} & \textbf{72.6/89.5} & \textbf{84.6/83.8}  & \textbf{81.56}\\
\hline 
\end{tabular}
\label{tableleaderboard:nonlin} 
\end{table*}

\begin{table*}[ht!]
\caption{BERT-Small results for Pro-KD on GLUE dev set} 
\small
\centering 
\begin{tabular}{c c c c c c c c c  c} 
\Xhline{4\arrayrulewidth} 
KD Method & CoLA & RTE & MRPC & STS-B & SST-2 & QNLI & QQP & MNLI  & Score\\ [0.5ex] 
\hline 
BERT$_\text{Large}$ & 61.89 & 68.96 & 88.22 & 89.58 & 92.89 & 92.4 & 90.23 & 86.1/86.25  & 83.79\\ 
\hline 
BERT-Small (NoKD) & 44.05&64.98 & 83.75 &	\textbf{87.41}&	88.3&	86.49&	88.43&	78.42/78.57&	77.74 \\
Vanilla KD & 43.28	&64.98	&84.96	&85.95	&88.65	&86.75&	88.24	&78.62	/78.55&	77.67
\\ 
TAKD & 43.79&	65.7&	83.98&	86.44&	88.88&	86.78&	88.4&	78.78/	78.64&	77.84\\
RCO&	44.32&	65.7	&84.91&	85.48 &	88.99 &	86.32 &	87.52 &	78.35	/78.85&	77.73\\
{Annealing KD} & \textbf{45}&	63.9&	87.09&	87.04&	89.56&	86.99&	88.58&	78.66/78.23 &	78.33 \\
\textbf{Pro-KD} & 42.37&	\textbf{66.79}&	\textbf{87.78}&{	87.09}&	\textbf{89.91}&	\textbf{87.88}&	\textbf{88.79}&	\textbf{79.18/	79.17}&	\textbf{78.72}.
\\
\hline 
\end{tabular}
\label{tablebertsmall:nonlin} 
\end{table*}

\begin{table*}[ht!]
\caption{The GLUE leaderboard test results of training BERT-Small from the BERT$_\text{Large}$ teacher.} 
\small
\centering 
\begin{tabular}{c c c c c c c c c  c} 
\Xhline{4\arrayrulewidth} 
KD Method & CoLA & RTE & MRPC & STS-B & SST-2 & QNLI & QQP & MNLI  & Score\\ [0.5ex] 
\hline 
BERT-Small (NoKD) & 41.3 &	62.6&	79.7&	80.05&	89&	86.3& 	78.05&	78.3/	77.6	& 74.37 \\
Vanilla KD & 37.3 &	63.4 &	80.55 & 	78.15	& 90.2& 	86.5&	78.25&	78/76.5&	73.95\\ 
TAKD & 38.5&	62.3&	80.5&	79.25&	89.7&	86.7&	78&	78.2/76.9&	74.06\\
RCO&	40.3 &	61.7&	79.75&	78.95&	90.6	&86.4&	78.35&	78.3/	77.3&	74.23
\\
{Annealing KD} &38.6&	63.1&	81.85&	80.6&	91.2&	87.3&	78.35&	77.8//	77.4	& 74.83
 \\
\textbf{Pro-KD} & 39 &	62.7	&82.9	&80.45&	91.2&	87.5&	79.15&	78.6	/78.2&	\textbf{75.16}
.
\\
\hline 
\end{tabular}
\label{tablebertsmall:nonlin} 
\end{table*}

\paragraph{Results}
As it is shown in Tables \ref{tableCV1:nonlin} and \ref{tableCV2:nonlin}, Pro-KD outperforms other baselines for both CIFAR-10 and CIFAR-100 experiments. Also, it is worth mentioning that the Annealing-KD technique is the second best result. RCO and TAKD both perform on-par with the regular KD and training the student from scratch.  

\begin{table}[h]
\caption{DistilRoBERTa results for Pro-KD on SQuAD} 
\centering 
\begin{tabular}{c c c} 
\Xhline{4\arrayrulewidth} 
KD Method & Squad 1.1 & Squad 2.0\\ [0.5ex] 
\hline 
Teacher & 93.7 & 87\\ 
\hline 
Vanilla KD & 85 & 73.65\\ 
TAKD & 85.4 & 73.8\\
\textbf{Pro-KD} & \textbf{86} & \textbf{76}\\
\hline 
\end{tabular}
\label{tablesquad1:nonlin} 
\end{table}

\begin{table}[h]
\caption{BERT-Small results for Pro-KD on SQuAD} 
\centering 
\begin{tabular}{c c c} 
\Xhline{4\arrayrulewidth} 
KD Method & Squad 1.1 & Squad 2.0\\ [0.5ex] 
\hline 
Teacher & 90.2 & 81\\ 
\hline 
Vanilla KD & 78.54 & 61.66\\ 
TAKD & 78.5 & 61.66\\
\textbf{Pro-KD} & \textbf{79.7} & \textbf{62.76}\\
\hline 
\end{tabular}
\label{tablesquad2:nonlin} 
\end{table}

\begin{table*}[h]
\small
\caption{Experimenting the impact of the adaptive temperature factor in Pro-KD using DistilRoBERTa trained with RoBERTa-large on the GLUE benchmark} 
\centering 
\begin{tabular}{c c c c c c c c c c c} 
\Xhline{4\arrayrulewidth} 
KD Method & CoLA & RTE & MRPC & STS-B & SST-2 & QNLI & QQP & MNLI (392K) & WNLI & Avg\\ [0.5ex]
         &    (8.5k) & (2.5k) &  (3.5k)   & (5.7k)  & (67k)      &  (108k)    & (363k)     & (392k)  &   &\\
\hline 
\hline
\textbf{Pro-KD} & \textbf{62.14} & \textbf{73.64} & \textbf{91.9} & \textbf{88.8} & \textbf{{92.66}} & 91.47 & 91.53 & 84.83/84.84 & \textbf{57.74} & \textbf{81.63}\\
\textbf{Pro-KD w/o $\mathcal{T}$} & 58.66 & 70.03 & 91.74 & 88.0 & 92.43 & \textbf{91.76} & \textbf{91.55} & \textbf{84.95/85.41} & 56.33 & 80.61\\

\hline 
\end{tabular}
\label{table1:ablation} 
\end{table*}

\subsection{Experimental Setup for Natural Language Understanding Tasks on the GLUE Benchmark}

\paragraph{Data}
We use the General Language Understanding Evaluation (GLUE) benchmark~\cite{wang-etal-2018-glue}, which consists of 9 natural language understanding tasks. The tasks cover textual entailment (RTE and MNLI), question-answer entailment (QNLI), paraphrase (MRPC), question paraphrase (QQP), sentiment (SST-2), textual similarity (STS-B), linguistic acceptability (CoLA), and Winograd Schema (WNLI).

\paragraph{Setup}
We perform experiments with multiple students of varying capacities. In the first experiment (Table \ref{tabledistilroberta:nonlin}), we use RoBERTa-large (24-layers)as teacher, DistilRoBERTa (6-layers) as student, and RoBERTa-base (12-layers) as the teacher-assistant for the TAKD baseline. For Pro-KD, we train the teacher for 5 epochs, and for training the student we use a maximum temperature of 5, increment factor of 2 for number of epochs, learning rate of 2e-5 and number of warmup epochs 2 for all tasks. For the second experiment (Table \ref{tablebertsmall:nonlin}), we use BERT-large (24-layers) as teacher, BERT-Small (4-layers) as student, and BERT-base (12-layers) as the teacher-assistant for TAKD. We train the teacher for 7 epochs, and for the student we use a maximum temperature of 7 for all tasks, increment factor of 1, and number of warmup epochs 1. For the learning rate, we use 5e-5 for RTE and MRPC, and 2e-5 for all other tasks. Additional details about other hyperparameters can be found in the appendix.

\paragraph{Results}
We present our results in Tables \ref{tabledistilroberta:nonlin}, \ref{tableleaderboard:nonlin} and \ref{tablebertsmall:nonlin}. Tables \ref{tabledistilroberta:nonlin} and \ref{tablebertsmall:nonlin} compare the performance of Pro-KD with Vanilla KD and TAKD on the GLUE dev set, while Table \ref{tableleaderboard:nonlin} presents the results of DistilRoBERTa on the test set based on GLUE benchmark's leaderboard. We see that Pro-KD significantly outperforms Vanilla KD and TAKD for both DistilRoBERTa and BERT-Small models. Even though TAKD is able to improve over Vanilla KD, the performance gain is much smaller compared to Pro-KD, demonstrating Pro-KD's high effectiveness in dealing with the large capacity gap problem.

\subsection{Experimental Setup for Question Answering}

\paragraph{Data}
We use the Stanford Question Answering Datasets (SQuAD v1.1 and SQuAD v2.0) which are a collection of 100k crowd-sourced question/answer pairs. Given a question and a passage from Wikipedia containing the answer, the task is to
predict the answer text span in the passage.The SQuAD 2.0 task extends the SQuAD 1.1 problem definition by allowing for the possibility that no short answer exists in the provided paragraph.

\paragraph{Setup}
Similar to the GLUE experiments, we perform experiments on 2 different students. In the first experiment (Table \ref{tablesquad1:nonlin}), we use RoBERTa-large (24-layers) as teacher, DistilRoBERTa (6-layers) as student, and RoBERTa-base (12-layers) as the teacher-assistant for the TAKD baseline. We train the teacher for 3 epochs, and for the student we use a maximum temperature of 3, increment factor of 1, number of warmup epochs 1, and a learning rate of 3e-5 with a batch size of 12 for both Squad v1.1 and Squad 2.0. For the second experiment (Table \ref{tablesquad2:nonlin}), we use BERT-large (24-layers) as teacher, BERT-Small (4-layers) as student, and BERT-base (12-layers) as the teacher-assistant for TAKD. We train the teacher for 3 epochs, and for the student we use a maximum temperature of 3, increment factor of 1, number of warmup epochs 1, and a learning rate of 3e-5 with a batch size of 12. Additional details about other hyperparameters can be found in the appendix.

\paragraph{Results}
We present our results in Tables \ref{tablesquad1:nonlin} and \ref{tablesquad2:nonlin}. We again see that Pro-KD consistently outperforms Vanilla KD and TAKD for both Squad v1.1 and Squad 2.0 tasks.

\subsection{Ablation Studies and Further Analysis}

\paragraph{The Impact of the Temperature Factor}

Here we show the impact of having the adaptive temperature factor in our technique. The adaptive temperature helps the student model to be exposed to a softened version of the teacher in earlier stages.
In this regard we repeated our experiment on the GLUE benchmark when the student is DistilRoBERTa and the teacher is RoBERTa-large. 
Table~\ref{table1:ablation} shows the result of Pro-KD with and without the temperature. The results indicate that dropping the adaptive temperature will hamper the performance of our model by about 1\% on the average GLUE score.

\section{Conclusion}
In this paper, we highlighted the importance of the capacity-gap problem in KD. 
We used the VC-dimension analysis to show that dealing with larger teachers may discount the benefit of using knowledge distillation in training. 
To address this capacity-gap problem, we introduced our Pro-KD technique. Our technique was based on defining a smoother training journey for the student by following the training footprints of the teacher instead of solely relying on distilling from a mature fully-trained teacher. In other words, in contrast to the regular knowledge distillation recipe for model compression in which the student model learns only from a fix fully-trained teacher, in our Pro-KD method, the student learns even from the training path of the teacher.  
We believe that following the training footsteps of the teacher can be quite influential in improving the student training and can lead to mitigating the capacity-gap problem in KD as well. 
We showed our theoretical analysis and justifications to support this idea; moreover, we evaluated our technique using a comprehensive set of experiments on different tasks such as image classification (CIFAR-10 and CIFAR-100), NLP language understanding tasks of the GLUE benchmark, and question answering (SQuAD 1.1 and 2.0) using BERT-based models and consistently got superior results.

\normalem
\bibliography{anthology,custom}
\bibliographystyle{acl_natbib}

\newpage
\appendix

\section{Why Does Pro-KD Work?}
\paragraph{Theoretical Justification}

In this section, first we state stopping the training procedure of the teacher at early epochs will improve Knowledge Distillation. Then, we argue that our method will exploit this fact without requiring any effort on finding the epoch at which early stopping the teacher leads to the optimum Knowledge Distillation. Finally, we will conclude that training the student together with the teacher leads to an important advantage that provides the student with the opportunity of exploiting the knowledge of the optimum teacher.  


In knowledge distillation, soft labels coming from a pre-trained teacher contains some so-called \textit{dark knowledge}~\cite{dong2019distillation} which gives a signal about the relative differences between the probabilities of the classes.  

It is shown by \cite{cho2019efficacy} and \cite{dong2019distillation} that early stopping during training of the teacher network improves the performance of Knowledge Distillation significantly; in other words, there is an optimal epoch at which if teacher training is stopped, the resulting teacher will act as a better teacher compared to the ones stopped at earlier or later epochs. 
Authors of \cite{cho2019efficacy} have shown this fact through experiments based on performing KD by using pre-trained teacher networks stopped at different epochs, and they also have indicated that changing the temperature cannot compensate lost Dark Knowledge. 
Authors of \cite{dong2019distillation} have provided mathematical justification showing that neural network learns more useful information faster. 
They have used the methods proposed in papers \cite{du2018gradient, oymak2019overparameterized, li2020gradient}, and the concept of Neural Tangent Kernel introduced by \cite{jacot2018neural} in order to reach an asymptotic conclusion. 
This conclusion states that for infinite wide neural networks, gradient descend algorithm searches over different direction with different pace; that is to say, the projection of the loss function in different eigenspaces evolves with different rate. Those rates can be calculated as follows \cite{dong2019distillation}:

\begin{equation}
\begin{matrix}
      \langle(u_t-y),e_i\rangle = \langle(I-\eta H^*)(u_t-y),e_i\rangle\\
      =\langle(u_t-y),(I-\eta H^*)e_i\rangle\\
      =(1-\eta \lambda_i)\langle(u_t-y),e_i\rangle
\end{matrix}
\end{equation}

where $u_t$ is the output of model, $y$ is the true label, $e_i$ is the eigenvector corresponding to the $i^{th}$ eigenvalue($\lambda_i$) of the static Gram matrix($H^*$) of the network. The gram matrix of neural networks is a function of time in general; however, in infinitely wide neural networks, this matrix will be static and called Neural Tangent Kernel \cite{jacot2018neural}. This static $n\times n$ matrix, n is the number of training samples, can be calculated as follows \cite{dong2019distillation}:

\begin{equation}
    H^* = \Big(\langle \frac{\partial f(\theta,x_i)}{\partial \theta}.\frac{\partial f(\theta,x_j)}{\partial \theta}\rangle \Big)_{i,j}
\end{equation}

where $f(\theta,\cdot)$ is the output of our model, $\theta$ are the parameters of the neural network, and $x_i$ is the $i^{th}$ data sample.

Arguments provided by \cite{dong2019distillation} state that a neural network learns more useful information faster than non-principal pieces of information about the input samples; however, it does not explain why continuing training teacher will decrease the level of that "Dark Knowledge". To justify this part, we use results from Tishby, Naftali, and Noga Zaslavsky. "Deep learning and the information bottleneck principle."\cite{tishby2015deep} and Saxe, Andrew M., et al. "On the information bottleneck theory of deep learning." \cite{saxe2019information}. In \cite{saxe2019information} and \cite{tishby2015deep}, authors show that there are two stages during training a deep neural network, initial fitting phase and compression phase. These two stages have some important characteristics:
\begin{enumerate}
    \item Initial Fitting: During this phase, the mutual information between the output of different layers and true labels is increasing. Also, the mutual information between output of layers and the input samples is increasing as well. In other words, network is gaining information about both labels and input samples. Therefore, network is gaining more aforementioned Dark Knowledge as it is gaining information about relative similarities between samples from different classes.
    \item Compression phase: During this phase, the mutual information between the output of different layers and true labels is increasing; however, the mutual information between output of layers and the input samples is decreasing. In other words, network tries to compress and discard information which it has gained about input samples \cite{saxe2019information}. We can state that Dark Knowledge is decreasing during this phase; that is because, network is forgetting relationships between data samples, and at the same time, it is gaining more information about labels. This procedure leads to more confidence on found probabilities for each class. As a result, the information about relative similarities between classes decreases and probabilities tend to one-hot vectors.  
\end{enumerate}

Based on the aforementioned facts, a teacher which has stopped learning at the optimum epoch provides more informative pieces of information to the student. Our method exploits the Dark Knowledge of the best teacher by training teacher and students together; in our method, during training teacher and student together, at some point teacher will have the highest level of Dark Knowledge which leads to the best student. The important point here is that we do not need to find the optimum epoch at which this highest level of Dark Knowledge will be achieved. Since we are performing Knowledge Distillation in fixed intervals and storing intermediate checkpoints, the student trained by guidance of the best teacher will be one of these checkpoints and we have access to that.

In contrast with usual KD methods, pre-trained teacher networks are not used in our methods; in fact, we train teacher alongside with the student. 
Based on the aforementioned arguments, Grow-KD outperforms other KD methods as it gives us the opportunity of exploiting the experience and the Dark Knowledge provided by the best teacher, the teacher which has stopped learning at the optimum epoch.

\section{Hyper-parameters}
In this section, we summarize the hyper-parameters used in our experiments.

\begin{table*}[h]
\caption{Model specific Hyper-parameters for BERT-Small on GLUE} 
\centering 
\begin{tabular}{c c c c c c c c c c} 
\Xhline{4\arrayrulewidth} 
Hyper-parameter & CoLA & RTE & MRPC & STS-B & SST-2 & QNLI & QQP & MNLI & WNLI\\ [0.5ex] 
\hline 
Learning Rate & 2e-5 & 5e-5 & 5e-5 & 2e-5 & 2e-5 & 2e-5 & 2e-5 & 2e-5 & 2e-5\\ 
N (Teacher Epochs) & 7 & 7 & 7 & 7 & 7 & 7 & 7 & 7 & 7\\
$\tau_{max}$ & 7 & 7 & 7 & 7 & 7 & 7 & 7 & 7 & 7\\
n (Phase 2 Epochs) & 10 & 10 & 10 & 10 & 10 & 10 & 10 & 10 & 10\\
\hline 
\end{tabular}
\label{table:nonlin} 
\end{table*}

\begin{table*}[h]
\caption{Common Hyper-parameters for DistilRoBERTa and BERT-Small models on GLUE} 
\centering 
\begin{tabular}{c c c c c c c c c c} 
\Xhline{4\arrayrulewidth} 
Hyper-parameter & CoLA & RTE & MRPC & STS-B & SST-2 & QNLI & QQP & MNLI & WNLI\\ [0.5ex] 
\hline 
Batch Size & 32 & 32 & 32 & 32 & 32 & 32 & 32 & 32 & 32\\ 
Max Seq. Length & 128 & 128 & 128 & 128 & 128 & 128 & 128 & 128 & 128\\
Vanilla KD Alpha & 0.5 & 0.5 & 0.5 & 0.5 & 0.5 & 0.5 & 0.5 & 0.5 & 0.5\\
Gradient Clipping & 1 & 1 & 1 & 1 & 1 & 1 & 1 & 1 & 1\\
\hline 
\end{tabular}
\label{table:nonlin} 
\end{table*}

\begin{table*}[h]
\caption{Model specific Hyper-parameters for DistilRoBERTa on GLUE} 
\centering 
\begin{tabular}{c c c c c c c c c c} 
\Xhline{4\arrayrulewidth} 
Hyper-parameter & CoLA & RTE & MRPC & STS-B & SST-2 & QNLI & QQP & MNLI & WNLI\\ [0.5ex] 
\hline 
Learning Rate & 2e-5 & 2e-5 & 2e-5 & 2e-5 & 2e-5 & 2e-5 & 2e-5 & 2e-5 & 2e-5\\ 
N (Teacher Epochs) & 5 & 5 & 5 & 5 & 5 & 5 & 5 & 5 & 5\\
$\tau_{max}$ & 5 & 5 & 5 & 5 & 5 & 5 & 5 & 5 & 5\\
n (Phase 2 epochs) & 10 & 10 & 10 & 10 & 10 & 10 & 10 & 10 & 10\\
\hline 
\end{tabular}
\label{table:nonlin} 
\end{table*}

\begin{table}[h]
\caption{Model specific Hyper-parameters for DistilRoBERTa on SQuAD} 
\centering 
\begin{tabular}{c c c c c c c c c c} 
\Xhline{4\arrayrulewidth} 
Hyper-parameter & SQuAD 1.1/2.0 & SQuAD Teacher\\ [0.5ex] 
\hline 
Learning Rate & 3e-5 & 1.5e-5\\ 
Batch Size & 12 & 12\\
Max Seq. Length & 384 & 384\\
Doc Stride & 128 & 128\\
Weight Decay & - & 0.01\\
N (Teacher Epochs) & 3 & 3\\
$\tau_{max}$ & 3 & -\\
n (Phase 2 Epochs) & 6 & -\\
\hline 
\end{tabular}
\label{table:nonlin} 
\end{table}

\begin{table}[ht!]
\caption{Model specific Hyper-parameters for BERT-Small on SQuAD} 
\centering 
\begin{tabular}{c c c c c c c c c c} 
\Xhline{4\arrayrulewidth} 
Hyper-parameter & SQuAD 1.1/2.0 & SQuAD Teacher\\ [0.5ex] 
\hline 
Learning Rate & 3e-5 & 3e-5\\ 
Batch Size & 12 & 12\\
Max Seq. Length & 384 & 384\\
Doc Stride & 128 & 128\\
N (Teacher Epochs) & 3 & 3\\
$\tau_{max}$ & 3 & -\\
n (Phase 2 Epochs) & 5 & -\\
\hline 
\end{tabular}
\label{table:nonlin} 
\end{table}





\end{document}